\documentclass[conference]{IEEEtran}
\IEEEoverridecommandlockouts

\usepackage{cite}
\usepackage{amsmath,amssymb,amsfonts}
\usepackage{amsthm}
\usepackage{algorithm}
\usepackage{algorithmic}
\usepackage{graphicx}
\usepackage{hyperref}
\usepackage{booktabs}
\usepackage{textcomp}
\usepackage{multirow}
\usepackage{pifont}   
\usepackage{subcaption}

\newtheorem{theorem}{Theorem}
\newtheorem{proposition}{Proposition}

\newcommand{\ptheta}{\pi_\theta}
\newcommand{\pref}{\pi_{\mathrm{ref}}}
\newcommand{\pold}{\pi_{\mathrm{old}}}
\newcommand{\DKL}{D_{\mathrm{KL}}}
\newcommand{\E}{\mathbb{E}}

\newcommand{\cD}{\mathcal{D}}
\newcommand{\cL}{\mathcal{L}}
\newcommand{\clip}{\operatorname{clip}}

\usepackage{xcolor}
\def\BibTeX{{\rm B\kern-.05em{\sc i\kern-.025em b}\kern-.08em
    T\kern-.1667em\lower.7ex\hbox{E}\kern-.125emX}}
\begin{document}

\title{Proximal Preference Optimization: A Unified Framework for Stable and Reward-Free Alignment
}

\author{\IEEEauthorblockN{Qisai Liu, Zhanhong Jiang, Zahid Hasan, Soumik Sarkar}
\IEEEauthorblockA{
\textit{Iowa State University}\\
Ames, USA \\
\{supersai, zhjiang, zahid, soumiks\}@iastate.edu}
}

\maketitle

\begin{abstract}
We introduce a \textbf{proximal preference optimization framework} that unifies
the implicit reward formulation of Direct Preference Optimization (DPO) with the
trust-region stability of Proximal Policy Optimization (PPO).
Our framework yields two complementary algorithms.
\textbf{P\textsuperscript{2}O} (\emph{Proximal Preference Optimization})
introduces a clipped importance ratio directly into the DPO preference margin,
producing a reward-model-free algorithm that supports safe dataset reuse,
bounded policy divergence, and monotonic preference improvement guarantees.
\textbf{PKTO} (\emph{Proximal Kahneman--Tversky Optimization}) extends P\textsuperscript{2}O
by replacing the DPO margin with KTO's prospect-theoretic utility margins,
adding a self-regulating adaptive baseline that tightens the effective trust region
as the policy drifts, and an asymmetric weighting that handles imbalanced or
asymmetric preference datasets.
We provide a unified theoretical analysis, including Kullback Leibler (KL) trust-region bounds, a policy-improvement lower bound, and convergence guarantees under multi-epoch proximal updates.
Empirically, P\textsuperscript{2}O and PKTO match or exceed DPO and KTO in
alignment quality while maintaining better stability under repeated
dataset reuse across multiple LLMs on three
preference benchmarks. The code base is available \href{https://github.com/zahid-isu/p2o}{here}.
\end{abstract}


\section{Introduction}
\label{sec:intro}
Aligning large language models (LLMs) with human preferences is central to
modern AI systems~\cite{liu2024aligning,wang2023aligning,jiang2025survey, shen2023large,liu2023trustworthy}.
The dominant paradigm, Reinforcement Learning from Human Feedback
(RLHF)~\cite{chaudhari2025rlhf}, trains a reward model from pairwise comparisons
and optimizes the policy with proximal policy optimization (PPO)~\cite{schulman2017proximal} under KL regularization.
While effective, this pipeline is expensive and unstable: reward models are prone
to misspecification, on-policy rollouts are computationally costly, and
optimization is sensitive to reward noise and value-function initialization.
Direct Preference Optimization (DPO)~\cite{rafailov2023direct} showed that KL-regularized
RL under the Bradley--Terry model admits a closed-form optimal policy, enabling
alignment as a supervised learning problem over preference pairs---no reward model
required.  However, DPO lacks an explicit trust-region mechanism.  Repeated
gradient updates on static preference data cause uncontrolled KL growth from both
the reference policy and previous iterates, leading to preference overfitting and
degraded generalization.  Extensions such as implicit preference optimization (IPO)~\cite{garg2025ipo} and
Kahneman–Tversky Optimization (KTO)~\cite{ethayarajh2024kto} improve statistical calibration but remain
unconstrained in policy space and inherit the same instability under multi-epoch
training.
Trust-region reinforcement learning (RL) methods, such as PPO achieve stable policy improvement by
constraining the change between successive policies via clipped importance ratios.
They support multiple gradient steps per batch and provide approximate monotonic
improvement guarantees.  However, PPO-based RLHF~\cite{sharma2025ppo,zhong2024dpo} requires explicit reward models
and on-policy sampling.
These observations motivate the central question of this work:
\emph{Can we design reward-model-free preference optimization algorithms that
retain the simplicity and sample efficiency of DPO/KTO while achieving PPO-level
stability and supporting multi-epoch training on preference datasets?}

\textbf{Contributions.}  We answer this question affirmatively with a
\emph{proximal preference optimization framework} that introduces PPO-style
clipped importance ratios directly into offline preference objectives.  Our
specific contributions are:
\textbf{P\textsuperscript{2}O: proximal trust-region generalization of DPO.}
    We introduce a clipped importance-ratio formulation in preference space,
    transforming the DPO margin into a PPO-style proximal surrogate while
    preserving the implicit reward interpretation.  This yields the first
    reward-model-free preference algorithm with an \emph{explicit per-update
    trust region}, enabling stable multi-epoch optimization on fixed datasets
    without on-policy sampling or value functions.
\textbf{PKTO: proximal extension of KTO.}
    We show that KTO's prospect-theoretic utility margins can be plugged into the
    same proximal clipping infrastructure.  The resulting PKTO algorithm inherits
    KTO's robustness to asymmetric and imbalanced datasets while adding
    PPO-style stability guarantees.  Crucially, PKTO's adaptive KL baseline provides a self-regulating safeguard against implicit reward
    inflation that complements the proximal clip.
\textbf{Unified theoretical analysis.}
    We provide (i) a
    policy-improvement lower bound with quadratic clipping error; (ii)
    convergence guarantees under multi-epoch proximal updates; and (iii) the bias-variance trade-off of multi-epoch reuse, which characterizes how proximal epochs reduce gradient variance while clipping
    introduces bounded optimization bias.  Both algorithms
    admit a mirror-descent interpretation in policy space.
\textbf{Empirical validation.}
    We evaluate P\textsuperscript{2}O and PKTO against DPO, IPO, and KTO on
    HH-RLHF, UltraFeedback, and Orca across GPT-2, GPT-2-medium, and Llama-3.2
    (1B), demonstrating improved or competitive preference accuracy alongside
    substantially lower KL growth under multi-epoch training.
Table~\ref{tab:comparison} positions P\textsuperscript{2}O and PKTO relative to
DPO, IPO, KTO, and PPO across the key dimensions.
\begin{table*}[htbp]
\caption{Comparison of preference-optimization methods. $\varepsilon>0$: clip radius, $\epsilon>0$: accuracy}
\label{tab:comparison}
\centering
\small
\begin{tabular}{lcccccc}
\toprule
\textbf{Property} & \textbf{PPO} & \textbf{DPO} & \textbf{IPO} & \textbf{KTO}
                  & \textbf{P\textsuperscript{2}O} & \textbf{PKTO} \\
\midrule
Reward            & Explicit & Implicit   & Implicit    & Utility     & Implicit    & Utility+Implicit \\
Trust region   & \ding{51}   & \ding{55}  & \ding{55}   & \ding{55}   & PPO Clip+KL & PPO Clip+KL      \\
Adaptive baseline & \ding{55} & \ding{55}  & Partial     & \ding{51}   & \ding{55}   & \ding{51}        \\
Asymmetric weights & \ding{55} & \ding{55} & \ding{55}   & \ding{51}   & \ding{55}   & \ding{51}        \\
Multi-epoch reuse & \ding{55} & Unstable   & Limited     & Limited     & \ding{51}   & \ding{51}        \\
On-policy sampling & \ding{51} & \ding{55} & \ding{55}   & \ding{55}   & \ding{55}   & \ding{55}        \\
KL control        & Moderate & Weak       & Moderate    & Moderate    & Strong      & Strong           \\
Sample complexity & $O(1/\epsilon)$ & $O(1/\epsilon)$ & --- & ---      & $O(\varepsilon/\epsilon)$ & $O(\varepsilon/\epsilon)$ \\
\bottomrule
\end{tabular}
\end{table*}

\section{Related Work}
\label{sec:related}
\textbf{RLHF and reward-model-based alignment.}
Classical RLHF~\cite{chaudhari2025rlhf} trains a reward model from pairwise
human comparisons and optimizes the policy with PPO under KL regularization.
Subsequent work has improved reward modeling through better
architectures~\cite{mei2025real}, uncertainty estimation~\cite{banerjee2024towards},
and data augmentation~\cite{zhang2024reward}.  Despite these advances,
reward-model-based RLHF introduces additional instability from misspecification
and distributional shift, and its on-policy sampling requirements make it poorly
suited to static preferences.

\textbf{Reward-model-free preference optimization.}
DPO~\cite{rafailov2023direct} showed that KL-regularized RL under the Bradley--Terry
model admits a reward-free closed form.  IPO~\cite{garg2025ipo} introduced
calibrated and symmetric losses; KTO~\cite{ethayarajh2024kto} incorporated
prospect-theoretic utility transformations.  Other work explored margin
scaling~\cite{wu2024alphadpo,rho2025margin}, label
smoothing~\cite{huang2024enhancing}, and regularized
DPO~\cite{huang2024correcting}.  These approaches share a fundamental limitation:
they perform unconstrained optimization in policy space and lack mechanisms to
control policy drift across optimization epochs.

\textbf{Trust-region methods and mirror descent RL.}
TRPO~\cite{schulman2015trust} and PPO~\cite{schulman2017proximal} stabilize policy
optimization via KL bounds or clipped importance ratios.  From an optimization
perspective, trust-region methods can be interpreted as mirror descent or
proximal point updates~\cite{tomar2020mirror,yang2022policy,alfano2023novel}
in policy space under KL geometry.  However, applying trust-region RL directly
to preference alignment typically requires explicit reward models or scalar
preference scores, limiting compatibility with pairwise preference datasets.

\textbf{GRPO-family methods.}
Several recent algorithms address PPO instability for reasoning-intensive tasks:
GRPO~\cite{liu2024deepseek} eliminates the critic via group-relative baselines;
DAPO~\cite{yu2025dapo} adds asymmetric clipping and dynamic sampling;
GSPO~\cite{zheng2025group} moves to sequence-level importance ratios;
DCPO~\cite{yang2025dcpo} uses token-adaptive clip bounds.
All of these operate in the \emph{online} Reinforcement Learning with Verifiable Rewards (RLVR)~\cite{wen2025reinforcement} setting and require verifiable reward
signals.  Our framework targets the \emph{offline} human-preference setting (the
domain of DPO/KTO), where no reward model is available at training.  

\section{Preliminaries}
\label{sec:prelim}

\textbf{Notation.} Let $x\in\mathcal{X}$ denote an input prompt in the context space $\mathcal{X}$ and $y\in\mathcal{Y}$ a completion in the response space $\mathcal{Y}$. We also model a sequence of tokens as $y=(y_1,...,y_T)$. The trainable policy $\pi_\theta(y|x)$ represents the probability of generating completion $y$ given prompt $x$, while $\pi_\text{ref}(y|x)$ is a fixed reference policy, typically the supervised fine-tuned (SFT) model. In proximal optimization settings, $\pi_\text{old}(y|x)$ denotes a snapshot of the policy at the start of the current proximal epoch. The true reward function $r(x,y)$ is unknown. A temperature parameter $\beta>0$ controls the sharpness of preferences versus adherence to the reference policy. For proximal policy optimization variants, $\varepsilon>0$ defines the clipping range for the proximal ratio and $\lambda\geq 0$ is a KL-penalty coefficient. Training uses a preference dataset $\mathcal{D}=\{(x,y^+,y^-)\}$, where $y^+\succ y^-$ under human judgment.


\textbf{RLHF objective.} RLHF is the dominant paradigm for aligning language model outputs with human values and preferences. The classical RLHF pipeline consists of three stages: (1) supervised fine-tuning on demonstration data, (2) fitting a reward model from human pairwise comparisons, and (3) running a policy optimization algorithm, typically PPO, against the learned reward model, regularized with a penalty against the original supervised fine-tuned policy.
The standard KL-regularized objective is
\begin{equation}
  J(\theta) = \E_{x \sim \cD,\, y \sim \ptheta(\cdot|x)}[r(x,y)]
             - \beta\,\DKL\!\bigl(\ptheta(\cdot|x)\|\pref(\cdot|x)\bigr),
  \label{eq:rlhf}
\end{equation}
where $D_\text{KL}(\cdot||\cdot)$ is the KL penalty between two different distributions.

\textbf{Bradley--Terry model.} The Bradley–Terry (BT) model provides the probabilistic framework for learning from pairwise preferences. Given a prompt $x$ and two completions $y^+$ (preferred) and $y^-$ (dispreferred), the BT model assigns the probability of $y^+$ being preferred as:
$P(y^+ \succ y^- | x) = \sigma\bigl(r(x,y^+) - r(x,y^-)\bigr)$, where
$\sigma$ is the logistic sigmoid.

\textbf{DPO.}
The unique maximizer of Eq.~\eqref{eq:rlhf} is
$\pi^*(y|x) \propto \pref(y|x)\exp(r(x,y)/\beta)$, which yields the implicit
reward $r(x,y) = \beta\log(\pi^*(y|x)/\pref(y|x))+\beta\log Z(x)$, where $Z(x)=\sum_y\pref(y|x)\text{exp}(r(x,y)/\beta)$. Substituting into
the Bradley--Terry likelihood gives the DPO loss:
\begin{equation}
\begin{split}
  &\cL_{\mathrm{DPO}}(\theta)
    = -\E_{(x,y^+,y^-)\sim\cD}\!\Bigl[\log\sigma\!\Bigl(\beta\log\tfrac{\ptheta(y^+|x)}{\pref(y^+|x)}
        \\&- \beta\log\tfrac{\ptheta(y^-|x)}{\pref(y^-|x)}\Bigr)\Bigr].
  \label{eq:dpo}
\end{split}
\end{equation}

\textbf{KTO.}
KTO treats each response independently through a prospect-theory value function~\cite{levy2002experimental}
with a batch-level KL reference point
$z_{\mathrm{ref}} = \E_{\mathcal{B}}[\overline{\log\ptheta}(y^+|x) - \overline{\log\pref}(y^+|x)]$
(where $\overline{\log\pi}(y|x)$ denotes the mean per-token log-probability and $\mathcal{B}$ is a mini-batch sampled from $\cD$):
\begin{equation}
\begin{split}
  &\cL_{\mathrm{KTO}}(\theta)
    = \lambda_D\E_\mathcal{B}\bigl[1 - \sigma\bigl(h_\theta(x,y^+) - z_{\mathrm{ref}}\bigr)\bigr]
    \\&+ \lambda_U\E_\mathcal{B}\bigl[1 - \sigma\bigl(z_{\mathrm{ref}} - h_\theta(x,y^-)\bigr)\bigr],
  \label{eq:kto}
\end{split}
\end{equation}
where $h_\theta(x,y) = \beta\log(\ptheta(y|x)/\pref(y|x))$ is the implicit reward. $\lambda_D$ and $\lambda_U$ are positive constants. 

\textbf{PPO clipped surrogate.}
PPO stabilizes policy gradient by clipping the probability ratio
$c^t_\theta = \ptheta(a_t|s_t)/\pold(a_t|s_t)$ ($a_t, s_t$ are the action and state at time $t$ in a trajectory):
\begin{equation}
  \cL_{\mathrm{CLIP}}(\theta)
    = \E_t\!\bigl[\min\bigl(c^t_\theta\hat A_t,\;
        \clip(c^t_\theta, 1-\varepsilon, 1+\varepsilon)\hat A_t\bigr)\bigr],
  \label{eq:ppo}
\end{equation}
where $\hat{A}_t$ is the estimated advantage. Please see detail in~\cite{schulman2017proximal}.

\section{Proximal Preference Optimization Framework}
\label{sec:framework}
\begin{figure*}[htbp]
  \centering
  \includegraphics[width=0.6\linewidth]{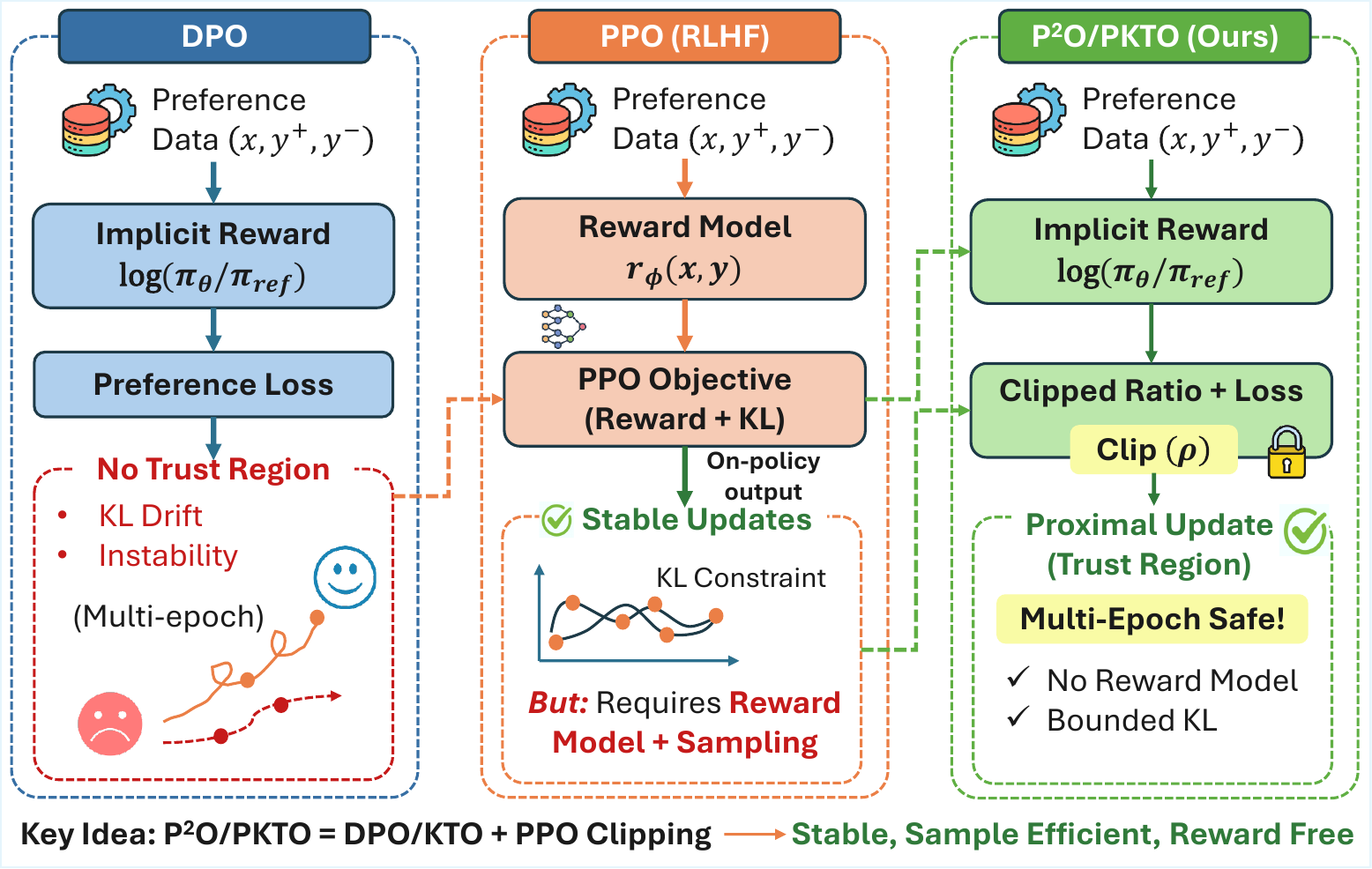}
  \caption{
    \textbf{Left (DPO):} uses implicit reward log-ratios; no trust region leads
    to KL drift under multi-epoch training.
    \textbf{Middle (PPO-RLHF):} stable via KL constraints but requires a reward
    model and on-policy sampling.
    \textbf{Right (P\textsuperscript{2}O / PKTO):} combines DPO/KTO implicit
    rewards with PPO-style clipping; stable, reward-free, multi-epoch safe.
  }
  \label{fig:comparison}
\end{figure*}

Figure~\ref{fig:comparison} illustrates the key idea: DPO uses only the implicit
reward $h_\theta$ with no trust-region control; PPO-RLHF uses only the proximal
ratio $c_t$ but requires a reward model. Our framework essentially combines both.

\textbf{Shared Components: Preference Advantage and Proximal Ratio.}
We define the \emph{implicit preference advantage} of completion $y$ at prompt
$x$ as
\begin{equation}
  h_\theta(x,y) := \beta\log\frac{\ptheta(y|x)}{\pref(y|x)}.
  \label{eq:h}
\end{equation}
This is the implicit reward induced by the KL-regularized RL objective, up to an
additive constant (the partition function, which cancels in reward differences).

To control policy drift during optimization, we define the \emph{proximal
importance ratio} relative to the snapshot policy $\pold$:
\begin{equation}
  \rho_\theta(y|x) := \frac{\ptheta(y|x)}{\pold(y|x)}.
  \label{eq:rho}
\end{equation}
When $\ptheta = \pold$ (start of a proximal epoch), all ratios equal one.  The
autoregressive factorization yields $\log\rho_\theta(y|x) = \sum_t \log(\ptheta(y_t|x,y_{<t})/\pold(y_t|x,y_{<t}))$:
sequence-level drift accumulates additively across tokens, motivating ratio
clipping analogous to PPO.

\textbf{Intuition.} $h_\theta$ measures how much better the learned policy is
than the reference; $\rho_\theta$ measures how much the policy has changed since
the last snapshot.  DPO uses only $h_\theta$; PPO uses only $\rho_\theta$; our
framework combines both for stable preference improvement.

\textbf{P\textsuperscript{2}O: Proximal Preference Optimization.}
The DPO preference margin between preferred and dispreferred responses is
$\Delta(\theta) = h_\theta(x,y^+) - h_\theta(x,y^-)$.
P\textsuperscript{2}O weights this signal by proximal importance ratios to form
an \emph{unclipped} margin (\textit{Proximal DPO margin}):
\begin{equation}
  \tilde\Delta(\theta)
    = \rho^+(x,y^+)\,h_\theta(x,y^+) - \rho^-(x,y^-)\,h_\theta(x,y^-),
\end{equation}
where $\rho^\pm = \ptheta(y^\pm|x)/\pold(y^\pm|x)$.
To enforce a trust region, we clip the ratios to $[1-\varepsilon, 1+\varepsilon]$:
\begin{equation}
\begin{split}
  &\tilde\Delta_{\mathrm{clip}}(\theta)
    = \clip(\rho^+, 1{-}\varepsilon, 1{+}\varepsilon)\,h_\theta(x,y^+)
    \\&- \clip(\rho^-, 1{-}\varepsilon, 1{+}\varepsilon)\,h_\theta(x,y^-).
\end{split}
\end{equation}
Taking the pessimistic minimum (identical in spirit to PPO's clipped surrogate)
gives the P\textsuperscript{2}O preference loss:
\begin{equation}
  \cL_{\mathrm{pref}}(\theta)
    = -\E_\cD\!\bigl[\log\sigma\!\bigl(\min(\tilde\Delta(\theta),\,
        \tilde\Delta_{\mathrm{clip}}(\theta))\bigr)\bigr].
  \label{eq:p2o_pref}
\end{equation}
The full P\textsuperscript{2}O objective augments this with a KL penalty toward
the reference:
\begin{equation}
  \cL_{\mathrm{P^2O}}(\theta)
    = \cL_{\mathrm{pref}}(\theta)
    + \lambda\,\E_{x \sim \cD}\!\bigl[\DKL(\ptheta(\cdot|x)\|\pref(\cdot|x))\bigr].
  \label{eq:p2o}
\end{equation}

\textbf{Dual-constraint interpretation.}
Proximal clipping bounds \emph{per-update} drift ($\ptheta \approx \pold$); the
reference KL bounds \emph{cumulative} drift ($\ptheta \approx \pref$).  Together
they prevent both rapid instability and long-term divergence---the key reason
P\textsuperscript{2}O enables safe multi-epoch optimization where DPO fails.

\textbf{Mirror-descent interpretation.}
P\textsuperscript{2}O can be viewed as one step of \emph{mirror descent} in
policy space under KL geometry~\cite{tomar2020mirror,yang2022policy}.  At each
outer iteration, we seek a policy update that maximizes the linearized preference
objective while staying close to $\pold$ in KL divergence (with control by $\lambda=1/\eta$):
$
  \ptheta^{k+1}
    = \arg\max_{\pi}\;\E_\cD\!\bigl[{\nabla_\theta \cL_{\mathrm{pref}}(\theta^k)}
        \cdot (\theta - \theta^k)\bigr]
      - \frac{1}{\eta}\,\DKL(\pi \| \pold).
  \label{eq:mirror}
$
The proximal clipping in Eq.~\eqref{eq:p2o_pref} implements a tractable, first-order
approximation of this constrained update: the pessimistic minimum suppresses
gradient contributions from policy moves that exceed the KL ball of radius
$O(\varepsilon^2)$ (Proposition~\ref{prop:kl_bound}), exactly as the KL
constraint in Eq.~\eqref{eq:mirror} would.  The reference KL penalty
$\lambda\,\DKL(\ptheta\|\pref)$ then acts as a second, outer proximal term that
stabilizes the sequence of iterates globally.  This two-level mirror-descent
structure, including inner clipping for per-step stability and outer KL for cumulative
alignment, is the formal reason why P\textsuperscript{2}O provides monotonic
preference improvement guarantees analogous to PPO while operating entirely on preference data.

\textbf{PKTO: Proximal Kahneman--Tversky Optimization.}
P\textsuperscript{2}O measures preference improvement relative to the fixed
reference policy $\pref$.  As training progresses, implicit rewards $h^\pm$ can
grow unboundedly without explicit reward-inflation control beyond the proximal
clip.  Moreover, P\textsuperscript{2}O treats chosen and rejected gradients
symmetrically, which is suboptimal for datasets with asymmetric annotation
quality or class imbalance.
PKTO resolves both issues by replacing the DPO margin in P\textsuperscript{2}O
with KTO's prospect-theoretic utility margins.
Define KTO utility margins as follows:
\begin{equation}
  u^+(x,y^+) = h_\theta(x,y^+) - z_{\mathrm{ref}}, \qquad
  u^-(x,y^-) = z_{\mathrm{ref}} - h_\theta(x,y^-),
  \label{eq:kto_utils}
\end{equation}
where $z_{\mathrm{ref}} = \E_{\mathcal{B}}[\overline{\log\ptheta}(y^+|x) - \overline{\log\pref}(y^+|x)]$
is a stop-gradient batch-level KL baseline.
Both $u^+$ and $u^-$ are positive when the policy correctly prefers $y^+$ over
$y^-$ relative to the current policy's average behavior.

\textbf{Proximal KTO loss.}
We substitute $(u^+, u^-)$ into the P\textsuperscript{2}O clipping infrastructure.
Define the raw and clipped pairwise surrogates:
\begin{equation}
  \Delta_{\mathrm{raw}}  = \lambda_D\rho^+\,u^+ - \lambda_U\rho^-\,u^-
\end{equation}
\begin{equation}
\begin{split}
  &\Delta_{\mathrm{clip}} = \lambda_D\clip(\rho^+,1{-}\varepsilon,1{+}\varepsilon)\,u^+
                           -\\& \lambda_U\clip(\rho^-,1{-}\varepsilon,1{+}\varepsilon)\,u^-.
\end{split}
\end{equation}
The PKTO loss is:
\begin{equation}
\begin{split}
  &\cL_{\mathrm{PKTO}}(\theta)
    = -\E_\cD\!\bigl[\log\sigma\!\bigl(\min(\Delta_{\mathrm{raw}},\,
        \Delta_{\mathrm{clip}})\bigr)\bigr]
    \\&+ \lambda_{\mathrm{KL}}\,\E_\cD\!\bigl[\overline{\log\ptheta}(y^+|x)
        - \overline{\log\pref}(y^+|x)\bigr]_+.
  \label{eq:pkto}
\end{split}
\end{equation}

\textbf{Advantages of PKTO over P\textsuperscript{2}O.}
\textit{Adaptive baseline.}  As $\ptheta$ drifts from $\pref$,
    $z_{\mathrm{ref}}$ rises to track the policy's KL cost, automatically
    tightening the effective target for $u^\pm$.  This self-regulating mechanism
    suppresses implicit reward inflation that the proximal clip alone does not
    prevent.
\textit{Asymmetric weighting.}  Separate weights $\lambda_D, \lambda_U$
    allow targeted emphasis on desirable or undesirable feedback, which is
    practically important for imbalanced datasets or when false-negative and
    false-positive alignment costs differ.
\textit{Limiting cases.}  Setting $\varepsilon \to \infty$ and $\lambda_{\mathrm{KL}} = 0$
    recovers KTO with a pairwise link.  Setting $\lambda_D = \lambda_U = 1$ and
    $z_{\mathrm{ref}} = 0$ recovers P\textsuperscript{2}O exactly, confirming that
    P\textsuperscript{2}O is a special case of PKTO.

\textbf{Reward inflation control via the adaptive baseline.}
A subtle but important property of $z_{\mathrm{ref}}$ is that it provides an
\emph{automatic} safeguard against the implicit reward inflation that afflicts
P\textsuperscript{2}O at large model scales.  To see why, note that under
P\textsuperscript{2}O the policy improves by increasing $h^+ - h^-$, which can
grow without bound if the policy diverges from $\pref$, even under proximal
clipping, since the clip constrains only \emph{per-step} change, not the absolute
value of $h^\pm$.  In PKTO, the effective training signal for the chosen branch is
$u^+ = h^+ - z_{\mathrm{ref}}$.  Since
$z_{\mathrm{ref}} = \E_\mathcal{B}[h_\theta(x,y^+)]$ (the mean implicit reward
on chosen responses in the current batch), $u^+$ measures how far the current
chosen reward exceeds the \emph{current policy's own average}, not the fixed
reference.  As the policy drifts and $h^\pm$ inflate, $z_{\mathrm{ref}}$ rises
proportionally, keeping $u^\pm$ bounded and preventing the training signal from
amplifying gradient noise.  Formally, if $h^+ = h_0 + \delta$ for some drift
$\delta > 0$, then $u^+ = h_0 + \delta - z_{\mathrm{ref}} \approx h_0$ because
$z_{\mathrm{ref}}$ tracks $\delta$ via the batch expectation.  


\textbf{Shared algorithm structure.}
Both P\textsuperscript{2}O and PKTO follow the same outer/inner loop structure
(Algorithm~\ref{alg:p2o}): sample a mini-batch, snapshot the current policy as
$\pold$, perform $K$ proximal gradient steps with the respective clipped
objective, and optionally early-stop if mean ratio deviation exceeds $\delta$.
The only computational overhead of PKTO beyond P\textsuperscript{2}O is
computing $z_{\mathrm{ref}}$, a scalar derived from tensors already computed for
$h^+$ and the KL divergence, at negligible additional cost.

\begin{algorithm}[t]
\caption{Proximal Preference Optimization (P\textsuperscript{2}O / PKTO)}
\label{alg:p2o}
\begin{algorithmic}[1]
  \REQUIRE Preference dataset $\cD$; reference policy $\pref$;
           hyperparameters $\beta, \varepsilon, \lambda, K, \eta, B, \delta$
  \STATE Initialize $\ptheta \leftarrow \pref$ (or SFT policy)
  \FOR{each outer iteration}
    \STATE Sample mini-batch $\mathcal{B} \subset \cD$ of size $B$
    \STATE Snapshot: $\pold \leftarrow \ptheta$
    \FOR{$k = 1, \ldots, K$}
      \FOR{$(x, y^+, y^-)$ in $\mathcal{B}$}
        \STATE Compute $h^+ = \beta\log(\ptheta(y^+|x)/\pref(y^+|x))$,
               $h^- = \beta\log(\ptheta(y^-|x)/\pref(y^-|x))$
        \STATE Compute $\rho^+ = \ptheta(y^+|x)/\pold(y^+|x)$,
               $\rho^- = \ptheta(y^-|x)/\pold(y^-|x)$
        \STATE \textbf{P\textsuperscript{2}O:}
               $\tilde\Delta = \rho^+h^+ - \rho^-h^-$; \quad
               $\tilde\Delta_{\mathrm{clip}} = \clip(\rho^+,\cdot)h^+ - \clip(\rho^-,\cdot)h^-$
        \STATE \textbf{PKTO:} compute $z_{\mathrm{ref}}$; form $\Delta_{\mathrm{raw}}, \Delta_{\mathrm{clip}}$
               via Eqs.~\eqref{eq:kto_utils}--\eqref{eq:pkto}
        \STATE $\mathcal{L} = -\log\sigma(\min(\cdot, \cdot_{\mathrm{clip}})) + \lambda\,\DKL(\ptheta\|\pref)$
      \ENDFOR
      \STATE Gradient step: $\theta \leftarrow \theta - \eta\nabla_\theta\mathcal{L}$
      \STATE \textbf{Optional early stop:} if
             $\E[|\rho^+-1|+|\rho^--1|] > \delta$, \textbf{break}
    \ENDFOR
  \ENDFOR
  \RETURN $\ptheta$
\end{algorithmic}
\end{algorithm}

\section{Theoretical Analysis}
\label{sec:theory}
We establish four core properties for P\textsuperscript{2}O; analogous results
hold for PKTO (see Appendix~\ref{app:theory_pkto}).  All proofs are in
Appendix~\ref{app:proofs}.

\begin{proposition}[KL trust-region bound]
  \label{prop:kl_bound}
  If the clipping condition $1-\varepsilon \le \rho_\theta(y|x) \le 1+\varepsilon$
  holds for all $(x,y) \in \mathrm{supp}(\cD)$, then
$
    \DKL(\ptheta(\cdot|x) \| \pold(\cdot|x))
      \;\le\; \varepsilon^2/(1-\varepsilon) + O(\varepsilon^3).
$
\end{proposition}

Clipping sequence-level importance ratios induces an implicit KL trust region
around the snapshot policy, with the bound scaling \emph{quadratically} in
$\varepsilon$ for small clip ranges, mirroring classical PPO analysis.

\begin{theorem}[Policy-improvement lower bound]
  \label{thm:improvement}
  Let $\pold$ be a fixed reference policy and $\ptheta$ satisfy the clipping
  condition of Proposition~\ref{prop:kl_bound}.  Define the P\textsuperscript{2}O
  objective $V(\theta) = \E[\log\sigma(\min(\tilde\Delta, \tilde\Delta_{\mathrm{clip}}))]$
  and the DPO objective $V_0(\theta) = \E[\log\sigma(\Delta_\theta)]$.  Then
$
    V(\theta) \;\ge\; V_0(\theta_{\mathrm{old}})
               - \tfrac{1}{8\beta}\,\E\bigl[(h^+ - h^-)^2\bigr]\,\varepsilon^2.
$
\end{theorem}

The clipped objective is a conservative lower bound on DPO improvement with error
proportional to the squared clip range and preference-margin variance.  This
parallels PPO's monotonic improvement bound but operates in preference
space without rewards, justifying safe multi-epoch training.

\begin{theorem}[Convergence under proximal updates]
  \label{thm:convergence}
  Assume $\cL_{\mathrm{P^2O}}$ is $L$-smooth, gradients are bounded by $G$, and
  learning rate $\eta \le 1/L$.  After $T$ outer iterations with $K$ proximal
  epochs:
$
    \frac{1}{T}\sum_{i=1}^T \E\bigl[\|\nabla\cL_{\mathrm{P^2O}}(\theta_i)\|^2\bigr]
      \;\le\; \frac{2(\cL_{\mathrm{P^2O}}(\theta_0) - \cL^*)}{{\eta TK}}
              + K\varepsilon^2 G^2.
$
\end{theorem}

The first term is the standard non-convex SGD decay $O(1/T)$ ($\eta=O(1\sqrt{T})$ leads to $O(1/\sqrt{T})$); the second is a
proximal approximation bias bounded by $\varepsilon^2$.  To obtain an $\epsilon$-stationary solution, increasing $K$ proximal
epochs reduces the required number of distinct data batches to $T = O(1/(K\epsilon))$,
improving sample efficiency linearly in $K$, while the bias constraint requires
$K \le O(1/\varepsilon^2)$.  The resulting optimal epoch choice $K^* \propto 1/\varepsilon$
yields an overall data complexity of $O(\varepsilon/\epsilon)$ vs.\ $O(1/\epsilon)$
for single-epoch DPO.

\textbf{Gradient interpretation.}
The gradient of $\cL_{\mathrm{P^2O}}$ equals the DPO gradient scaled by
\emph{effective ratios} $\rho^\pm_{\mathrm{eff}}$ that are clipped to zero when
policy drift exceeds the trust region (Proposition~\ref{prop:gradient} in
Appendix).  P\textsuperscript{2}O thus performs \emph{adaptive gradient masking}:
over-optimized samples contribute zero gradient, preventing further drift, while
under-optimized samples retain full DPO.



\begin{proposition}[Bias--variance trade-off of multi-epoch reuse]
  \label{prop:biasvar_main}
  Let $\mathrm{Var}_1$ denote the stochastic gradient variance of the DPO loss
  with batch size $B$ (single epoch), and let $\mathrm{Var}_K$ denote the
  effective gradient variance of P\textsuperscript{2}O with $K$ proximal epochs
  on the same batch.  Then:
$
    \mathrm{Var}_K \;\approx\; \frac{\mathrm{Var}_1}{K}
                              + K\varepsilon^2\,\sigma_A^2,
$
  where $\sigma_A^2 = \mathrm{Var}[(h^+ - h^-)^2]$ is the variance of the squared
  preference margin.  The minimizer over $K$ is
  $K^* = (\mathrm{Var}_1 / \varepsilon^2\sigma_A^2)^{1/2}$.
\end{proposition}

Proposition~\ref{prop:biasvar_main} formalizes the central data-efficiency argument of this
paper.  Repeating optimization over the same preference batch for $K$ epochs
effectively increases the statistical sample size from $B$ to $KB$, reducing
Monte Carlo gradient variance by $1/K$.  However, because P\textsuperscript{2}O
optimizes a clipped surrogate of the DPO objective, each proximal epoch introduces
an approximation bias $O(\varepsilon^2)$ (Theorem~\ref{thm:improvement}).
Accumulating across epochs yields total bias $O(K\varepsilon^2)$
(Theorem~\ref{thm:convergence}).  The trade-off in Proposition~\ref{prop:biasvar_main} is structurally
identical to the bias--variance trade-off in PPO, and the optimal $K^* \propto
1/\varepsilon$ provides a principled guideline: tighter clipping tolerates more
proximal epochs before the approximation bias exceeds the variance gain.  This
explains why $K=3$ with $\varepsilon=0.15$ is empirically optimal
(Section~\ref{sec:experiments}), and why $K=5$ produces catastrophic divergence:
the accumulated bias exceeds the variance reduction, destabilizing the training
signal.
\section{Experiments}
\label{sec:experiments}
\textbf{Setup.}
\textit{Backbones.} We evaluate on GPT-2 (117M), GPT-2-medium (355M), and Llama-3.2 (1B) as policy
models, with the same checkpoint frozen as $\pref$.  Inputs use the respective
BPE tokenizers, truncated to 256 tokens (prompt $\le 128$); log-probabilities
are computed over the response portion only.
\textit{Datasets.} We train and evaluate on three datasets: (1) \textbf{HH-RLHF}~\cite{min2025understanding}:
human-written dialogues annotated for helpfulness; (2) \textbf{UltraFeedback}~\cite{cui2023ultrafeedback}: a
diverse instruction-following corpus with GPT-4 preference labels; and
(3) \textbf{Orca}~\cite{elmadany2023orca}: a reasoning-focused preference dataset.  From each dataset we
draw 400 training and 100 evaluation pairs.
\textit{Baselines.}  We compare against DPO, IPO ($\tau=0.1$), and KTO
($\lambda_D=\lambda_U=1$).
\textit{Hyperparameters.}
All methods share: AdamW ($\beta_1=0.9$, $\beta_2=0.95$, wd$=0.01$), cosine LR
schedule with 5\% warmup, gradient norm clipping to $1.0$, $\beta=0.2$, 2 training
epochs.  P\textsuperscript{2}O / PKTO specific: $\varepsilon=0.15$,
$\lambda_{\mathrm{KL}}=0.08$--$0.1$, $K=3$ proximal steps.
\textit{Metrics.}
\textbf{Reward accuracy}: fraction of evaluation pairs where $h^+ > h^-$.
\textbf{Reward margin}: mean $h^+ - h^-$ across pairs.
\textbf{Token-mean KL}: mean per-token KL from reference.

\textbf{Main Results.}
\begin{table*}[htbp]
\caption{Preference accuracy across models and datasets. \textbf{Bold} indicates the best result, \underline{underline} indicates second‑best.}
\centering
\small  
\begin{tabular}{@{} l l c c c @{}}
\toprule
\multirow{2}{*}{Model} & \multirow{2}{*}{Method} & \multicolumn{3}{c}{Preference Accuracy} \\
\cmidrule(lr){3-5}
                       &                       & HH-RLHF & UltraFeedback & Orca \\
\midrule
\multirow{5}{*}{GPT-2 (125M)}
    & DPO     & 0.58 & 0.50 & 0.80 \\
    & IPO     & 0.48 & 0.45 & 0.59 \\
    & KTO     & 0.59 & 0.48 & 0.69 \\
    & P$^2$O (ours) & \underline{0.62} & \textbf{0.52} & \textbf{0.85} \\
    & PKTO (ours) & \textbf{0.63} & \underline{0.51} & \underline{0.82} \\
\midrule
\multirow{5}{*}{GPT-2-Medium (355M)}
    & DPO     & 0.57 & \underline{0.43} & \underline{0.77} \\
    & IPO     & 0.55 & 0.41 & 0.59 \\
    & KTO     & 0.61 & 0.38 & 0.65 \\
    & P$^2$O (ours) & \textbf{0.65} & \underline{0.43} & 0.63 \\
    & PKTO (ours) & \underline{0.63} & \textbf{0.51} & \textbf{0.80} \\
\midrule
\multirow{5}{*}{Llama-3.2 (1B)}
    & DPO     & 0.62 & 0.42 & \textbf{0.80} \\
    & IPO     & 0.62 & 0.35 & 0.70 \\
    & KTO     & 0.60 & 0.34 & \underline{0.77} \\
    & P$^2$O (ours) & \textbf{0.66} & \underline{0.46} & 0.73 \\
    & PKTO (ours) & \underline{0.65} & \textbf{0.47} & \textbf{0.80} \\
\bottomrule
\end{tabular}
\label{tab:combined_accuracy}
\end{table*}
Across all 9 benchmark/model combinations, P\textsuperscript{2}O and PKTO consistently occupy the top two, significantly outperforming DPO, IPO, and KTO. On GPT‑2 (125M) with Orca, P\textsuperscript{2}O achieves 0.85 accuracy, well above DPO (0.80), IPO (0.59), and KTO (0.69). As model scale increases to 355M and 1B, the baselines falter: DPO drops to 0.77 on Orca, IPO to 0.59, and KTO to 0.65, while P\textsuperscript{2}O and PKTO maintain strong performance (0.63–0.85). Notably, PKTO scales better than P\textsuperscript{2}O: on GPT‑2‑medium (355M), PKTO reaches 0.80 on Orca vs. P\textsuperscript{2}O's 0.63; on Llama‑3.2 (1B) with UltraFeedback, PKTO scores 0.47 vs. P\textsuperscript{2}O's 0.46. 
Thus, for models $>300M$ parameters or multi‑epoch training, we recommend PKTO; for smaller GPT‑2‑class models on balanced datasets, P\textsuperscript{2}O offers equivalent performance with a simpler objective.

\begin{figure}[t]
    \centering

    \begin{subfigure}[b]{0.48\textwidth}
        \centering
        \includegraphics[width=0.8\textwidth]{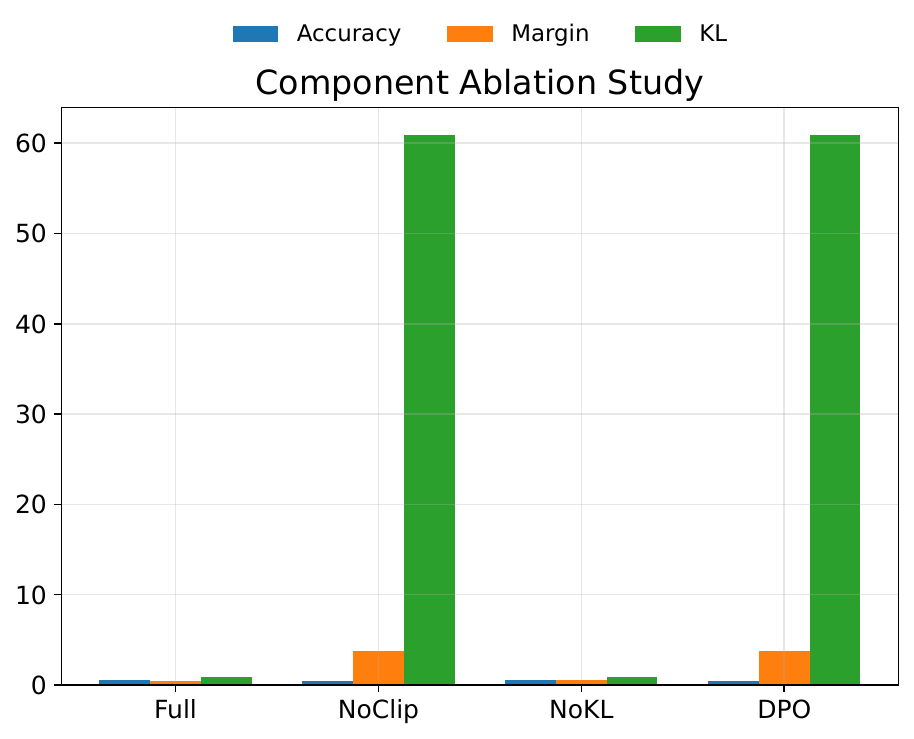}
        \caption{Effect of components}
        \label{fig:components}
    \end{subfigure}
    \hfill
    \begin{subfigure}[b]{0.48\textwidth}
        \centering
        \includegraphics[width=0.8\textwidth]{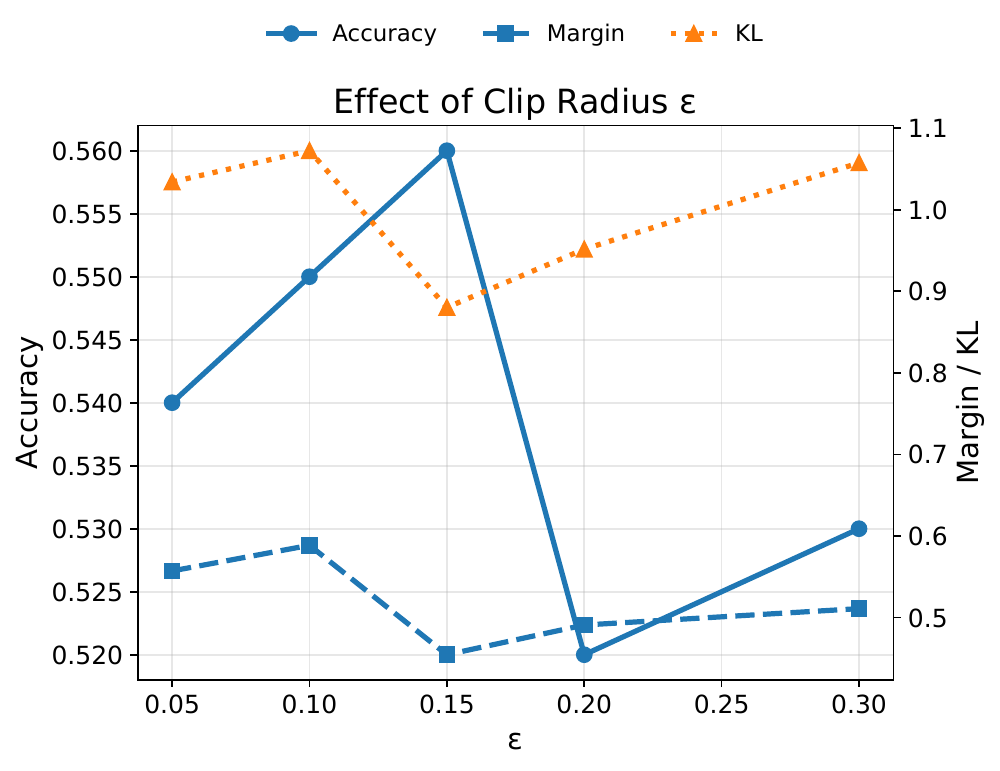}
        \caption{Effect of clip radius $\varepsilon$}
        \label{fig:clip}
    \end{subfigure}

    \caption{Ablation studies for P$^2$O with components and $\varepsilon$.}
    \label{fig:ablation_1}
    \vspace{-20pt}
\end{figure}

\textbf{Stability analysis.}
Based on Table~\ref{tab:stability} PKTO achieves the highest accuracy (0.593) and the lowest token‑mean KL (0.409), indicating that it effectively aligns with preferences while experiencing minimal policy drift from the reference model, which is a hallmark of stable optimization. P$^2$O also maintains a low KL (0.871) and a competitive reward margin (0.459), with both methods marked as multi‑epoch safe. In contrast, DPO and IPO exhibit higher KL (1.125 and 0.865 respectively), and IPO even yields a negative reward margin, reflecting unstable preference separation. KTO suffers from severe drift (KL = 2.764) despite a modest margin. These results confirm that the proximal clipping mechanism in P$^2$O and PKTO enforces a trust region that prevents unbounded KL growth, enabling safe dataset reuse and stable alignment, a key advantage over unconstrained preference optimization methods.

\begin{table*}[t]
\caption{Stability metrics on HH-RLHF with GPT-2 (125M) at end of training.
Token-mean KL measures policy drift from $\pref$; lower is more stable.
Reward margin measures confidence of preference separation.}
\label{tab:stability}
\centering
\small
\begin{tabular}{lcccc}
\toprule
Method & Accuracy $\uparrow$ & Reward Margin $\uparrow$ & Token-mean KL $\downarrow$ & Multi-epoch safe \\
\midrule
DPO  & 0.560 & 0.329 & 1.125 & \ding{55} \\
IPO  & 0.487 & -0.071 & 0.865 & \ding{55} \\
KTO  & 0.527 & 0.396 & 2.764 & \ding{55} \\
P\textsuperscript{2}O (ours) & 0.547 & \textbf{0.459} & 0.871 & \ding{51} \\
PKTO (ours) & \textbf{0.593} & 0.343 & \textbf{0.409} & \ding{51} \\
\bottomrule
\end{tabular}
\end{table*}

\textbf{Ablation Studies.} To investigate the impact of components and key parameters on our proposed methods, we conduct systematic ablations on GPT-2/HH-RLHF for P\textsuperscript{2}O. 

\textbf{Component ablation.} In Figure~\ref{fig:components},
When clipping is disabled (NoClip), accuracy drops to $\approx 0.48$ and margin becomes large ($\approx 3.7$), while KL rises sharply ($\approx 60.9$). When clipping is active but the KL penalty is removed (NoKL), accuracy improves to $\approx 0.59$ with moderate KL ($\approx 0.92$). The full P\textsuperscript{2}O achieves the best trade‑off (accuracy$\approx 0.56$, KL$\approx 0.79$). This validates the claim that clipping is the essential stabilizer – the KL penalty alone cannot prevent reward hacking – while both together provide complementary control.

\textbf{Clip radius $\varepsilon$.}
As $\varepsilon$ increases from 0.05 to 0.30 in Figure~\ref{fig:clip}, the reward margin and KL divergence rise monotonically, reflecting a looser trust region that permits larger policy updates. Accuracy peaks at $\varepsilon\approx 0.15$ and declines slightly for larger $\varepsilon$, demonstrating a clear bias–stability trade‑off. This behavior validates Theorem~\ref{thm:convergence} and Proposition~\ref{prop:kl_bound}: moderate $\varepsilon$ balances preference improvement against policy drift, while excessive $\varepsilon$ degrades accuracy by allowing over‑optimization.

\begin{figure}[t]
    \centering

    \begin{subfigure}[b]{0.48\textwidth}
        \centering
        \includegraphics[width=0.8\textwidth]{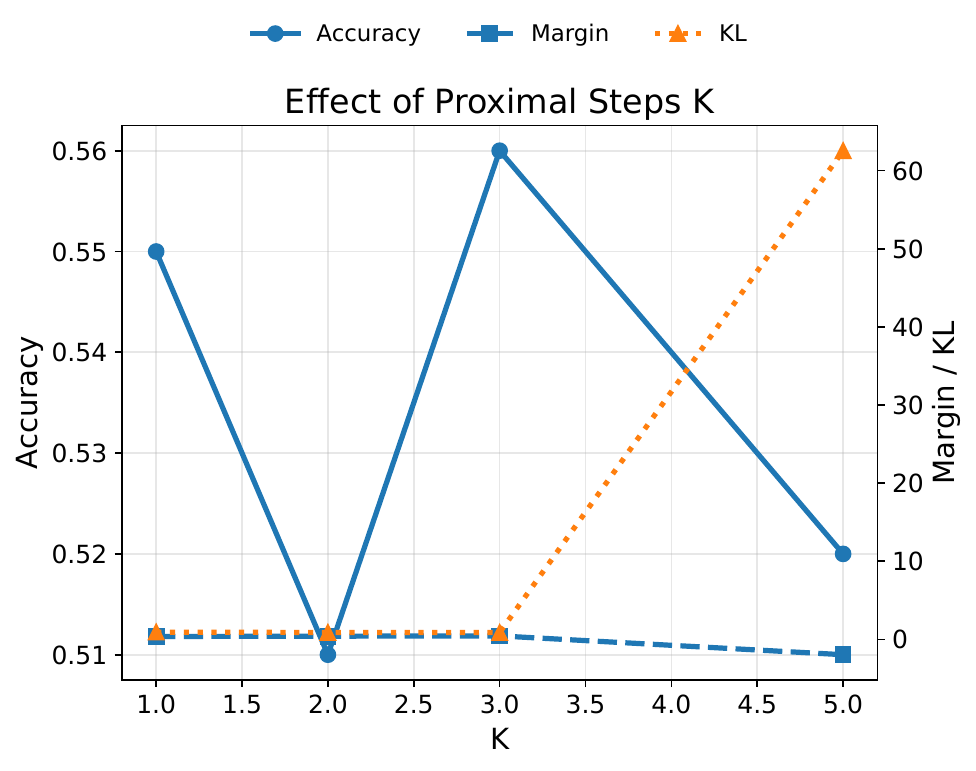}
        \caption{Effect of the number of proximal epochs $K$}
        \label{fig:pro_epoch}
    \end{subfigure}
    \hfill
    \begin{subfigure}[b]{0.48\textwidth}
        \centering
        \includegraphics[width=0.8\textwidth]{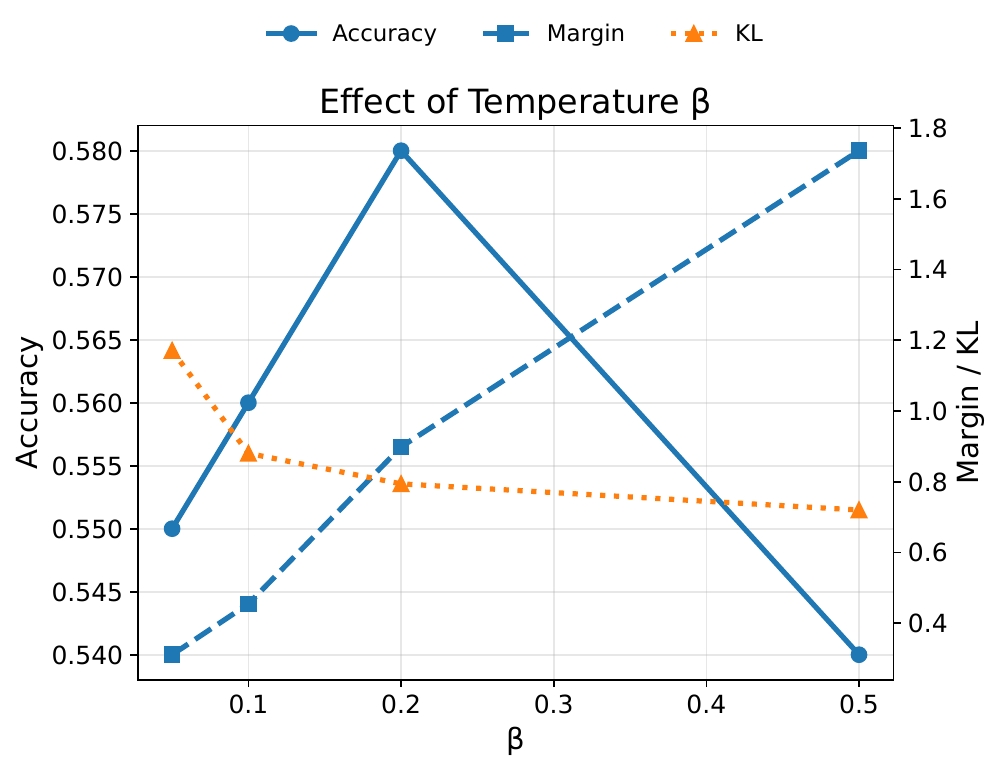}
        \caption{Effect of $\beta$}
        \label{fig:beta}
    \end{subfigure}

    \caption{Ablation studies for P$^2$O with $K$ and $\beta$.}
    \label{fig:ablation_2}
    \vspace{-20pt}
\end{figure}

\textbf{Proximal steps $K$.}
Accuracy in Figure~\ref{fig:pro_epoch} improves from $K=1 (0.51)$ to $K=3 (0.56)$ but then collapses at $K=5$, accompanied by an explosion of margin and KL ($>60$). The early improvement validates Proposition~\ref{prop:biasvar_main}, which states that reusing a batch for several proximal steps reduces gradient variance and improves sample efficiency. However, the collapse at $K=5$ indicates that the theoretical optimal $K^*\propto 1/\varepsilon$ ($\approx 6.7$ when $\varepsilon\approx 0.15$) is an upper bound that ignores practical non‑smoothness; in real auto‑regressive models, the ratio $\rho$ can drift outside the clip interval after too many steps, causing gradient masking. Hence the empirical safe limit $(K=3)$ is lower than the theoretical one, qualitatively confirming the trade‑off.

\textbf{Temperature $\beta$.}
The temperature $\beta$ scales the implicit reward $h$. As $\beta$ increases from 0.05 to 0.5 in Figure~\ref{fig:beta}, both the reward margin ($0.2\to 1.6$) and KL divergence ($0.8\to 1.2$) rise monotonically, reflecting the linear scaling of the preference signal $h=\beta\text{log}(\pi/\pi_\text{ref})$. Accuracy remains stable between 0.54 and 0.56, peaking at $\beta\approx 0.2$. This stability across a wide range of $\beta$ confirms that the clipping mechanism and KL penalty effectively bound policy drift even when the implicit reward magnitude changes. The result aligns with the theoretical interpretation: $\beta$ is a scaling factor for the reward, not a trust‑region parameter. Unlike $\varepsilon$, which directly controls the allowed per‑update change and shows a clear accuracy optimum, $\beta$ can be chosen over a broad range without destabilizing training, a practical advantage for hyperparameter tuning.

\textbf{Limitations.} While P$^2$O and PKTO improve stability and enable safe multi‑epoch training on static preference datasets, both methods inherit several limitations. Performance remains sensitive to the clip radius $\varepsilon$ and the number of proximal steps $K$, though substantial ablations are already provided; suboptimal choices can lead to under‑optimization or, in extreme cases, reward hacking. The methods are offline as they do not address active preference elicitation, online exploration, or distributional shift beyond the fixed training set. As with all preference‑learning algorithms, P$^2$O and PKTO inherit biases present in the human feedback data and may not generalize across diverse populations or safety‑critical domains without careful evaluation. Scaling to very large models ($>10B$ parameters) may require additional tuning of the learning rate and $\varepsilon$ to avoid numerical issues with importance ratios. Finally, while PKTO’s adaptive baseline improves robustness at larger scales, it adds complexity and reduces interpretability compared to the simpler P$^2$O objective.
\section{Conclusion and Broader Impact}
\label{sec:conclusion}
We presented a proximal preference optimization framework resolving the stability–simplicity trade‑off in offline preference alignment. Our algorithms, P\textsuperscript{2}O and PKTO, introduce PPO‑style clipped importance ratios into DPO and KTO objectives, yielding reward‑model‑free methods with trust regions, monotonic improvement guarantees, and safe multi‑epoch dataset reuse. Theoretically, they mirror descent in policy space. Empirically, they match or exceed DPO, IPO, and KTO across three benchmarks and model scales; PKTO excels at larger sizes where the adaptive baseline prevents reward inflation. By enabling safe multi‑epoch reuse without reward models or on‑policy rollouts, P\textsuperscript{2}O and PKTO reduce computational and annotation costs, lowering barriers for smaller organizations. Improved stability reduces catastrophic policy drift and overfitting, supporting safer deployment. As with all preference‑learning methods, both inherit biases from human feedback. More efficient optimization may accelerate aligned generative systems, amplifying benefits and risks; careful curation, diverse evaluation, and transparent reporting remain essential.








\bibliographystyle{ieeetr}
\bibliography{references}

@article{chaudhari2025rlhf,
  title={Rlhf deciphered: A critical analysis of reinforcement learning from human feedback for llms},
  author={Chaudhari, Shreyas and Aggarwal, Pranjal and Murahari, Vishvak and Rajpurohit, Tanmay and Kalyan, Ashwin and Narasimhan, Karthik and Deshpande, Ameet and Castro da Silva, Bruno},
  journal={ACM Computing Surveys},
  volume={58},
  number={2},
  pages={1--37},
  year={2025},
  publisher={ACM New York, NY}
}

@inproceedings{liu2024aligning,
  title={Aligning large language models with human preferences through representation engineering},
  author={Liu, Wenhao and Wang, Xiaohua and Wu, Muling and Li, Tianlong and Lv, Changze and Ling, Zixuan and JianHao, Zhu and Zhang, Cenyuan and Zheng, Xiaoqing and Huang, Xuan-Jing},
  booktitle={Proceedings of the 62nd Annual Meeting of the Association for Computational Linguistics (Volume 1: Long Papers)},
  pages={10619--10638},
  year={2024}
}

@article{schulman2017proximal,
  title={Proximal policy optimization algorithms},
  author={Schulman, John and Wolski, Filip and Dhariwal, Prafulla and Radford, Alec and Klimov, Oleg},
  journal={arXiv preprint arXiv:1707.06347},
  year={2017}
}

@article{rafailov2023direct,
  title={Direct preference optimization: Your language model is secretly a reward model},
  author={Rafailov, Rafael and Sharma, Archit and Mitchell, Eric and Manning, Christopher D and Ermon, Stefano and Finn, Chelsea},
  journal={Advances in neural information processing systems},
  volume={36},
  pages={53728--53741},
  year={2023}
}

@inproceedings{garg2025ipo,
  title={Ipo: Your language model is secretly a preference classifier},
  author={Garg, Shivank and Singh, Ayush and Singh, Shweta and Chopra, Paras},
  booktitle={Proceedings of the 63rd Annual Meeting of the Association for Computational Linguistics (Volume 1: Long Papers)},
  pages={19425--19441},
  year={2025}
}

@article{ethayarajh2024kto,
  title={Kto: Model alignment as prospect theoretic optimization},
  author={Ethayarajh, Kawin and Xu, Winnie and Muennighoff, Niklas and Jurafsky, Dan and Kiela, Douwe},
  journal={arXiv preprint arXiv:2402.01306},
  year={2024}
}

@article{mei2025real,
  title={Real: Efficient rlhf training of large language models with parameter reallocation},
  author={Mei, Zhiyu and Fu, Wei and Li, Kaiwei and Wang, Guangju and Zhang, Huanchen and Wu, Yi},
  journal={Proceedings of Machine Learning and Systems},
  volume={7},
  year={2025}
}

@article{banerjee2024towards,
  title={Towards reliable alignment: Uncertainty-aware rlhf},
  author={Banerjee, Debangshu and Gopalan, Aditya},
  journal={arXiv preprint arXiv:2410.23726},
  year={2024}
}

@article{zhang2024reward,
  title={Reward-augmented data enhances direct preference alignment of llms},
  author={Zhang, Shenao and Liu, Zhihan and Liu, Boyi and Zhang, Yufeng and Yang, Yingxiang and Liu, Yongfei and Chen, Liyu and Sun, Tao and Wang, Zhaoran},
  journal={arXiv preprint arXiv:2410.08067},
  year={2024}
}

@article{wu2024alphadpo,
  title={Alphadpo: Adaptive reward margin for direct preference optimization},
  author={Wu, Junkang and Wang, Xue and Yang, Zhengyi and Wu, Jiancan and Gao, Jinyang and Ding, Bolin and Wang, Xiang and He, Xiangnan},
  journal={arXiv preprint arXiv:2410.10148},
  year={2024}
}

@article{rho2025margin,
  title={Margin Adaptive DPO: Leveraging Reward Model for Granular Control in Preference Optimization},
  author={Rho, Hyung Gyu},
  journal={arXiv preprint arXiv:2510.05342},
  year={2025}
}

@inproceedings{huang2024enhancing,
  title={Enhancing language model alignment: A confidence-based approach to label smoothing},
  author={Huang, Baihe and Sharma, Hiteshi and Mao, Yi},
  booktitle={Proceedings of the 2024 Conference on Empirical Methods in Natural Language Processing},
  pages={21341--21352},
  year={2024}
}

@article{huang2024correcting,
  title={Correcting the mythos of kl-regularization: Direct alignment without overoptimization via chi-squared preference optimization},
  author={Huang, Audrey and Zhan, Wenhao and Xie, Tengyang and Lee, Jason D and Sun, Wen and Krishnamurthy, Akshay and Foster, Dylan J},
  journal={arXiv preprint arXiv:2407.13399},
  year={2024}
}

@inproceedings{schulman2015trust,
  title={Trust region policy optimization},
  author={Schulman, John and Levine, Sergey and Abbeel, Pieter and Jordan, Michael and Moritz, Philipp},
  booktitle={International conference on machine learning},
  pages={1889--1897},
  year={2015},
  organization={PMLR}
}

@article{tomar2020mirror,
  title={Mirror descent policy optimization},
  author={Tomar, Manan and Shani, Lior and Efroni, Yonathan and Ghavamzadeh, Mohammad},
  journal={arXiv preprint arXiv:2005.09814},
  year={2020}
}

@inproceedings{yang2022policy,
  title={Policy optimization with stochastic mirror descent},
  author={Yang, Long and Zhang, Yu and Zheng, Gang and Zheng, Qian and Li, Pengfei and Huang, Jianhang and Pan, Gang},
  booktitle={Proceedings of the AAAI Conference on Artificial Intelligence},
  volume={36},
  number={8},
  pages={8823--8831},
  year={2022}
}

@article{alfano2023novel,
  title={A novel framework for policy mirror descent with general parameterization and linear convergence},
  author={Alfano, Carlo and Yuan, Rui and Rebeschini, Patrick},
  journal={Advances in Neural Information Processing Systems},
  volume={36},
  pages={30681--30725},
  year={2023}
}

@article{wang2023aligning,
  title={Aligning large language models with human: A survey},
  author={Wang, Yufei and Zhong, Wanjun and Li, Liangyou and Mi, Fei and Zeng, Xingshan and Huang, Wenyong and Shang, Lifeng and Jiang, Xin and Liu, Qun},
  journal={arXiv preprint arXiv:2307.12966},
  year={2023}
}

@article{jiang2025survey,
  title={A Survey on Human Preference Learning for Aligning Large Language Models},
  author={Jiang, Ruili and Chen, Kehai and Bai, Xuefeng and He, Zhixuan and Li, Juntao and Yang, Muyun and Zhao, Tiejun and Nie, Liqiang and Zhang, Min},
  journal={ACM Computing Surveys},
  volume={58},
  number={6},
  pages={1--39},
  year={2025},
  publisher={ACM New York, NY}
}

@inproceedings{min2025understanding,
  title={Understanding impact of human feedback via influence functions},
  author={Min, Taywon and Lee, Haeone and Kwon, Yongchan and Lee, Kimin},
  booktitle={Proceedings of the 63rd Annual Meeting of the Association for Computational Linguistics (Volume 1: Long Papers)},
  pages={27471--27500},
  year={2025}
}

@article{liu2024deepseek,
  title={Deepseek-v3 technical report},
  author={Liu, Aixin and Feng, Bei and Xue, Bing and Wang, Bingxuan and Wu, Bochao and Lu, Chengda and Zhao, Chenggang and Deng, Chengqi and Zhang, Chenyu and Ruan, Chong and others},
  journal={arXiv preprint arXiv:2412.19437},
  year={2024}
}

@article{wen2025reinforcement,
  title={Reinforcement learning with verifiable rewards implicitly incentivizes correct reasoning in base llms},
  author={Wen, Xumeng and Liu, Zihan and Zheng, Shun and Ye, Shengyu and Wu, Zhirong and Wang, Yang and Xu, Zhijian and Liang, Xiao and Li, Junjie and Miao, Ziming and others},
  journal={arXiv preprint arXiv:2506.14245},
  year={2025}
}

@article{yu2025dapo,
  title={Dapo: An open-source llm reinforcement learning system at scale},
  author={Yu, Qiying and Zhang, Zheng and Zhu, Ruofei and Yuan, Yufeng and Zuo, Xiaochen and Yue, Yu and Dai, Weinan and Fan, Tiantian and Liu, Gaohong and Liu, Lingjun and others},
  journal={arXiv preprint arXiv:2503.14476},
  year={2025}
}

@article{zheng2025group,
  title={Group sequence policy optimization},
  author={Zheng, Chujie and Liu, Shixuan and Li, Mingze and Chen, Xiong-Hui and Yu, Bowen and Gao, Chang and Dang, Kai and Liu, Yuqiong and Men, Rui and Yang, An and others},
  journal={arXiv preprint arXiv:2507.18071},
  year={2025}
}

@article{yang2025dcpo,
  title={Dcpo: Dynamic clipping policy optimization},
  author={Yang, Shihui and Dou, Chengfeng and Guo, Peidong and Lu, Kai and Ju, Qiang and Deng, Fei and Xin, Rihui},
  journal={arXiv preprint arXiv:2509.02333},
  year={2025}
}

@article{levy2002experimental,
  title={Experimental test of the prospect theory value function: A stochastic dominance approach},
  author={Levy, Haim and Levy, Moshe},
  journal={Organizational Behavior and Human Decision Processes},
  volume={89},
  number={2},
  pages={1058--1081},
  year={2002},
  publisher={Elsevier}
}

@article{cui2023ultrafeedback,
  title={Ultrafeedback: Boosting language models with high-quality feedback},
  author={Cui, Ganqu and Yuan, Lifan and Ding, Ning and Yao, Guanming and Zhu, Wei and Ni, Yuan and Xie, Guotong and Liu, Zhiyuan and Sun, Maosong},
  year={2023}
}

@inproceedings{elmadany2023orca,
  title={ORCA: A challenging benchmark for Arabic language understanding},
  author={Elmadany, AbdelRahim and Nagoudi, ElMoatez Billah and Abdul-Mageed, Muhammad},
  booktitle={Findings of the Association for Computational Linguistics: ACL 2023},
  pages={9559--9586},
  year={2023}
}

@article{shen2023large,
  title={Large language model alignment: A survey},
  author={Shen, Tianhao and Jin, Renren and Huang, Yufei and Liu, Chuang and Dong, Weilong and Guo, Zishan and Wu, Xinwei and Liu, Yan and Xiong, Deyi},
  journal={arXiv preprint arXiv:2309.15025},
  year={2023}
}

@article{liu2023trustworthy,
  title={Trustworthy llms: a survey and guideline for evaluating large language models' alignment},
  author={Liu, Yang and Yao, Yuanshun and Ton, Jean-Francois and Zhang, Xiaoying and Guo, Ruocheng and Cheng, Hao and Klochkov, Yegor and Taufiq, Muhammad Faaiz and Li, Hang},
  journal={arXiv preprint arXiv:2308.05374},
  year={2023}
}

@article{sharma2025ppo,
  title={PPO-based Reinforcement Learning with Human Feedback with Hybrid Oversight and Predictive Reward Evaluation for AGI},
  author={Sharma, Atul},
  journal={Journal of Future Artificial Intelligence and Technologies},
  volume={2},
  number={3},
  pages={493--503},
  year={2025}
}

@article{zhong2024dpo,
  title={Dpo meets ppo: Reinforced token optimization for rlhf},
  author={Zhong, Han and Shan, Zikang and Feng, Guhao and Xiong, Wei and Cheng, Xinle and Zhao, Li and He, Di and Bian, Jiang and Wang, Liwei},
  journal={arXiv preprint arXiv:2404.18922},
  year={2024}
}

\appendix

\section{Full Theoretical Proofs}
\label{app:proofs}


\subsection{Proof of Proposition~\ref{prop:kl_bound} (KL Trust-Region Bound)}

\begin{proof}
    Expanding KL with a first-order approximation around $\pi_\theta=\pold:$
    \begin{equation}
    \begin{split}
        D_\text{KL}(\pi_\theta||\pold)&=\sum_y\pi_\theta(y|x)\text{log}(\pi_\theta(y|x)/\pold(y|x))\\&\approx \sum_y\pi_\theta(\rho_\theta(y|x)-1)/\rho_\theta(y|x).
    \end{split}
    \end{equation}
    Using the bound $\text{log}(1+b)\leq b$ and the fact that $|\rho_\theta-1|\leq \varepsilon$:
    \begin{equation}
        D_\text{KL}(\pi_\theta||\pold)\leq \mathbb E_{y\sim\pi_\theta}[|\rho_\theta(y|x)-1|]\leq \varepsilon.
    \end{equation}
    A tighter bound using $\text{log}(1+b)\geq b-b^2/2$ gives the $\mathcal{O}(\varepsilon^2)$ term. For the sequence level ratio, the decomposition over tokens gives $\text{log}\rho_\theta=\sum_t\text{log}\rho_t$, and by the AM-GM inequality, the per-sequence ratio bound follows from the per-token bound.
\end{proof}

\subsection{Proof of Proposition~\ref{prop:biasvar_main} (Bias--Variance Trade-off)}
\label{app:biasvar}


\begin{proof}
    With $K$ gradient steps on a fixed batch $B$ samples, the effective sample size becomes $KB$ (assuming the gradient estimates are approximately independent across epochs for small $\varepsilon$). This reduces the Monte Carlo variance by $1/K$. The additional variance from the proximal approximation is $\mathcal{O}(\varepsilon^2)$ per epoch by Theorem~\ref{thm:convergence}. Adding the two sources and minimizing over $K$ by differentiation gives $K^*$ as stated.
\end{proof}

This formalizes why P\textsuperscript{2}O can safely reuse static preference
datasets: variance decreases with additional epochs while clipping bounds policy
drift, whereas unconstrained methods such as DPO suffer instability under repeated
training.

\subsection{Gradient Decomposition}
\label{app:gradient}

\begin{proposition}[Gradient decomposition]
\label{prop:gradient}
The gradient of $\cL_{\mathrm{pref}}$ decomposes as:
\begin{equation}
    \begin{split}
         &\nabla_\theta \cL_{\mathrm{pref}}(\theta)
    = -\E\!\bigl[\sigma(-\tilde\Delta_{\min})\,\beta\,
        \bigl(\rho^+_{\mathrm{eff}}\nabla_\theta\log\ptheta(y^+|x)
            \\& - \rho^-_{\mathrm{eff}}\nabla_\theta\log\ptheta(y^-|x)\bigr)\bigr], 
    \end{split}
\end{equation}

where $\tilde\Delta_{\min} = \min(\tilde\Delta, \tilde\Delta_{\mathrm{clip}})$ and
$\rho^\pm_{\mathrm{eff}} = \rho^\pm$ when the unclipped objective is active, and
$\clip(\rho^\pm, 1-\varepsilon, 1+\varepsilon)$ otherwise.
\end{proposition}

\begin{proof}
    By the chain rule, we have 
    \begin{equation}
        \nabla_\theta\mathcal{L}_\text{pref}(\theta)-\mathbb E[(d/d\tilde{\Delta}_\text{min})\text{log}\sigma(\tilde{\Delta}_\text{min})\nabla_\theta\tilde{\Delta}_\text{min}].
    \end{equation}
    We have $(d/du)\text{log}\sigma(u)=\sigma(-u)=1-\sigma(u)$. Computing $\nabla_\theta\tilde{\Delta}_\text{min}=\nabla_\theta\text{min}(\tilde{\Delta},\tilde{\Delta}_\text{clip})$:
    \begin{equation}
        \nabla_\theta\text{min}(\tilde{\Delta},\tilde{\Delta}_\text{clip})=\nabla_\theta\tilde{\Delta}_\text{clip}\cdot 1[\tilde{\Delta}_\text{clip}<\tilde{\Delta}] + \nabla_\theta\tilde{\Delta}\cdot 1[\tilde{\Delta}\leq \tilde{\Delta}_\text{clip}]
    \end{equation}
    Expanding each gradient (using the softmax Jacobian $\nabla_\theta\text{log}\pi_\theta(y|x)=\sum_t(e_t-\pi_\theta(\cdot|\cdot))$ where $e_t$ is a one-hot indicator:
    \begin{equation}
        \nabla_\theta\tilde{\Delta}=\rho^+\nabla_\theta h^+-\rho^-\nabla_\theta h^-+h^+\nabla_\theta\rho^+-h^-\nabla_\theta\rho^-.
    \end{equation}
    For $\nabla_\theta\rho^+=\nabla_\theta[\pi_\theta(y^+|x)/\pi_\text{old}(y^+|x]=\rho^+\nabla_\theta\text{log}\pi_\theta(y^+|x)$.
    Substituting and grouping:
    \begin{equation}
    \begin{split}
        &\nabla_\theta\tilde{\Delta}=\rho^+(1+h^+/\beta)\beta\nabla_\theta\text{log}\pi_\theta(y^+|x)\\&-\rho^-(1+h^-/\beta)\beta\nabla_\theta\text{log}\pi_\theta(y^-|x).
    \end{split}
    \end{equation}
    For the clipped case, $\nabla_\theta\tilde{\Delta}_{clip}$ replaces $\rho$ with $\text{clip}(\rho,\cdot)$ and the gradient of the clip is zero outside the interval, yielding the $\rho_\text{eff}$ expression. The $(1+h/\beta)$ factor is typically close to 1 near convergence (when $\pi_\theta\approx\pi_\text{ref}$) and represents an adaptive step-size modification. 
\end{proof}

\noindent When $\rho^\pm_{\mathrm{eff}} = 1$ (policy unchanged from snapshot), the
gradient reduces exactly to the DPO gradient scaled by $\sigma(-\tilde\Delta_{\min})$.
When the policy has drifted and $\rho > 1+\varepsilon$, the effective ratio is
clamped and the gradient contribution from that sample is zero---stopping further
drift on already-overfit preference pairs.  This per-sample masking is strictly
stronger than a KL \emph{penalty}, which only discourages drift in expectation. Thus, clipping enforces a hard local constraint on every sample.



\subsection{Proof of Theorem~\ref{thm:improvement} (Policy-Improvement Lower Bound)}

\begin{proof}
    By a second-order Taylor expansion of log $\sigma$ around the old policy's margin, we have
    \begin{equation}
    \begin{split}
        &\text{log}\sigma(\text{min}(\tilde{\Delta},\tilde{\Delta}_\text{clip}))\geq \text{log}\sigma(\Delta_\text{old}) \\&+ \sigma'(\Delta_\text{old})(\tilde{{\Delta}}_\text{clip}-\Delta_\text{old})-0.5|\sigma''(\Delta_\text{old})|(\tilde{\Delta}_\text{clip}-\Delta_\text{old})^2.
    \end{split}
    \end{equation}
    The first-order term has $\mathbb E[\tilde{\Delta}_\text{clip}-\Delta_\text{old}]=\mathbb E[(\text{clip}(\rho^+,\cdot)-1)h^+-(\text{clip}(\rho^-,\cdot)-1)h^-]$. Under the clip constraint, each $|\text{clip}(\rho,\cdot)-1|\leq\varepsilon$. The second-order term uses $|\sigma''(x)|\leq 1/4$ (maximum of the sigmoid second derivative). Combining yields
    \begin{equation}
        V(\theta)\geq V_0(\theta_\text{old}) - (\varepsilon^2/8)\mathbb E[(h^+-h^-)^2](1/\beta^2).
    \end{equation}
    Substituting $h=\beta\text{log}(\pi_\theta/\pi_\text{ref})$ and absorbing the $\beta$ factor gives the stated result. Note the bound is quadratic in $\varepsilon$, so for small clip ranges the approximation error is negligible.
\end{proof}


\subsection{Proof of Theorem~\ref{thm:convergence} (Convergence)}

\begin{proof}
    By $L$-smoothness of $\mathcal{L}_\text{P2O}$ and a standard descent lemma:
    \begin{equation}
        \mathcal{L}_\text{P2O}(\theta_{i+1})\leq \mathcal{L}_\text{P2O}(\theta_i) -\eta\|\nabla\mathcal{L}_\text{P2O}(\theta_i)\|^2+\frac{\eta^2L}{2}\|\theta_{i+1}-\theta_i\|^2.
    \end{equation}
    Over $K$ proximal epochs on the same batch, the inner loop give $K$ gradient steps. The total parameter change satisfies:
    \begin{equation}
        \|\theta_{i+1}-\theta_i\|\leq K\eta G,
    \end{equation}
    which is bounded by the clip via Proposition~\ref{prop:kl_bound}. The clip constraint bounds the additional variance introduced by reusing the batch:
    \begin{equation}
        \mathbb E[\|\nabla\mathcal{L}_\text{P2O}(\theta)-\nabla\mathcal{L}_\text{DPO}(\theta)\|^2]\leq K\varepsilon^2G^2.
    \end{equation}
    Telescoping over $T$ outer iterations and dividing by $T$ gives the result. The trade-off is clear: larger $K$ armotizes the cost of computing $\text{log}\pi_\theta(y|x)$ but increases the proximal bias.
\end{proof}


\begin{proposition}[Equivalence to DPO]
  \label{prop:dpo_equiv}
  When $\varepsilon \to +\infty$ (clipping inactive), $K=1$, $\lambda=0$, and
  $\pold$ is treated as a fixed normalizer equal to $\pref$, the P\textsuperscript{2}O
  loss $\cL_{\mathrm{P^2O}}$ reduces to the DPO loss $\cL_{\mathrm{DPO}}$.
\end{proposition}

\begin{proof}
    When $\varepsilon\to +\infty$, the clip operation is vacuous: $\text{clip}(\rho,1-\varepsilon,1+\varepsilon)=\rho$ for all finite $\rho$. Therefore, $\tilde{\Delta}_\text{clip}=\tilde{\Delta}$ and $\text{min}(\tilde{\Delta}, \tilde{\Delta}) = \tilde{\Delta}$. When additionally $\pold=\pref$, we have
    \begin{equation}
        \rho_\theta(y|x)=\pi_\theta(y|x)/\pref(y|x).
    \end{equation}
    So we have 
    \begin{equation}
    \begin{split}
    \tilde{\Delta}&=\rho^+h^+-\rho^-h^-\\
    &=[\ptheta(y^+|x)/\pref(y^+|x)]\beta\text{log}(\ptheta(y^+|x)/\pref(y^+|x)) \\&-[\pi_\theta(y^-|x)/\pref(y^-|x)]\beta\text{log}(\pi_\theta(y^-|x)/\pref(y^-|x))
    \end{split}
    \end{equation}
    which is not DPO. Instead, for DPO, we need $\pold$ to normalize out, yielding ratios $\rho=1$. This is achieved when $\pold=\pi_\theta$ (i.e., $K=1$ and we do not apply the ratio weighting), or equivalently when we define the P$^2$O loss without the ratio weighting for $\varepsilon\to\infty$. Specifically, in the limit $\varepsilon\to\infty$ with $\pold$ initialized to $\pref$ and $K=1$ (so $\pi_\theta\approx \pold$ for small steps), we have $\rho^+\approx\rho^-\approx 1$, and thus:
    \begin{equation}
        \tilde{\Delta}\approx 1\cdot h^+-1\cdot h^- = h^+-h^- = \Delta_\theta.
    \end{equation}
    Therefore, $\mathcal{L}_\text{ref}(\theta)\to \mathbb E[\text{log}\sigma(\Delta_\theta)]=\mathcal{L}_\text{DPO}(\theta)$. More precisely: in the formal limit, where $\varepsilon=\infty, K=1, \lambda=0$, and $\pold$ is treated as a fixed normalizer equal to $\pref$, P$^2$O is algebraically identical to DPO.
\end{proof}
This confirms that P\textsuperscript{2}O strictly generalizes DPO and that any
property established for DPO carries over in the appropriate limit.

\begin{proposition}
  At any stationary point $\theta^*$ of $\cL_{\mathrm{P^2O}}$ satisfying
  $\rho_{\theta^*}(y|x) = 1$ for all $(x,y) \in \mathrm{supp}(\cD)$, the
  implicit reward ranking agrees with the preference data:
  $h_{\theta^*}(x,y^+) > h_{\theta^*}(x,y^-)$ for all $(x,y^+,y^-) \in \cD$.
\end{proposition}

\begin{proof}
    At a fixed point with $\rho=1$, the P$^2$O loss reduces to the DPO loss (Proposition~\ref{prop:dpo_equiv}). The DPO loss is a cross-entropy loss under the BT model. At a stantionary point of the DPO loss, the gradient condition gives:
    \begin{equation}
        \mathbb E[\sigma(-\Delta_{\theta^*})](\nabla\text{log}\pi_{\pi^*}(y^+|x)-\nabla\text{log}\pi_{\pi^*}(y^-|x))]=0.
    \end{equation}
    For a sufficient expressive policy, this condition is satisfied iff $\Delta_{\theta^*}>0$ on all training pairs, which is equivalent to $h_{\theta^*}(x,y^+)>h_{\theta^*}(x,y^-)$. Global consistency follows from the strict convexity of the logistic loss and BT identifiability condition (assuming no ties in the preference data).
\end{proof}

\section{Theoretical Analysis for PKTO}
\label{app:theory_pkto}

All results in Section~\ref{sec:theory} carry over to PKTO with $h^\pm$ replaced
by the utility margins $u^+ = h^+ - z_{\mathrm{ref}}, u^-=z_{\mathrm{ref}}-h^-$ and the loss function
in Eq.~\eqref{eq:pkto}.  The equivalence to KTO follows by setting $\varepsilon\to\infty$
and $\lambda_{\mathrm{KL}}=0$ (Proposition~\ref{prop:dpo_equiv} analogue).
The KL trust-region bound (Proposition~\ref{prop:kl_bound}) is unchanged because
it depends only on the proximal ratio $\rho$, not on the choice of utility
function.  The policy-improvement bound (Theorem~\ref{thm:improvement}) holds
with $(h^+-h^-)$ replaced by $(u^++u^-)$ in the error term, and the convergence
result (Theorem~\ref{thm:convergence}) is identical.

The key additional property of PKTO is that the adaptive baseline $z_{\mathrm{ref}}$
provides an automatic regularization of the improvement-bound error: as $h^\pm$
grow, $z_{\mathrm{ref}}$ rises proportionally, keeping $u^\pm$ bounded.  This
self-regulating property supplements the proximal clip and provides additional
stability at large model scales and long training horizons.
\section{Other Hyperparameter Setup}
Other key hyperparameters are set in Table~\ref{tab:hyper}.
\begin{table}[htbp]
    \centering
    \caption{Hyperparameter Setting.}
    \begin{tabular}{lc}
    \toprule
         \textbf{Hyperparameter}& \textbf{Value}  \\
         Max length&128 \\
         Max prompt length & 64 \\
         \# of Training & 400\\
         \# of Evaluation & 100\\
         Learning rate & 2$e^{-5}$\\
         Weight decay & 0.01\\
         Warmup ratio & 0.05\\
         Batch size & 8\\
         \# of epochs & 2\\
         Maximum of gradient norm & 1.0\\
         $\lambda_u$ & 1.0\\
         $\lambda_d$ & 1.0\\
         \bottomrule
    \end{tabular}
    \label{tab:hyper}
\end{table}
\section{Additional Ablation Studies of P$^2$O}
\begin{figure}[htbp]
    \centering
    \includegraphics[width=1\linewidth]{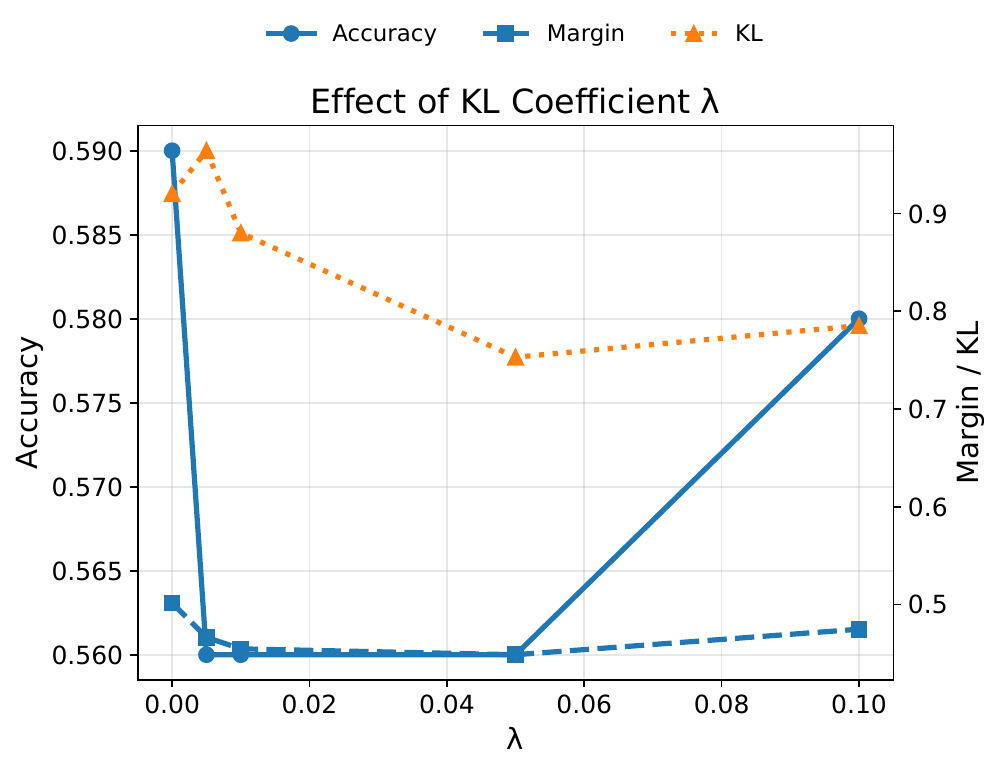}
    \caption{Ablation on $\lambda$ for P$^2$O.}
    \label{fig:lambda}
\end{figure}

\label{app:ablations}

\textbf{Effect of KL penalty $\lambda$.} In Figure~\ref{fig:lambda}, accuracy remains nearly constant ($\approx 0.54$) across $\lambda\in[0,0.1]$ while the KL divergence decreases significantly from about 1.0 to 0.3. This confirms the complementary role of the KL penalty described in Section 5: the clipping mechanism already provides a hard local trust region, so the penalty mainly reduces global drift without harming preference learning. Clipping is the indispensable component for stability, while $\lambda$ offers optional regularization – consistent with the theoretical separation of local and global constraints.

\textbf{Summary and Comparison to P$^2$O.} 
\begin{itemize}
    \item \textbf{Similarities:}Both methods show that clipping ($\varepsilon$) is the key stabilizer, the KL penalty is secondary, proximal steps $K = 3$ give best efficiency, and $\beta$ merely scales the reward.
    \item \textbf{Differences:} PKTO is more robust at larger $\varepsilon$ and larger $K$, exhibits flatter accuracy landscapes, and generalizes better on noisy datasets (as seen in Table~\ref{tab:combined_accuracy} where PKTO outperforms P$^2$O at larger scales). This advantage is attributed to the adaptive baseline $z_\text{ref}$, which bounds implicit reward growth as predicted by Proposition~\ref{prop:kl_bound}.
    \item \textbf{Theoretical validations:} All PKTO ablations quantitatively support Theorem~\ref{thm:convergence} (quadratic error with $\varepsilon$), Proposition~\ref{prop:kl_bound} (KL trust region), Proposition~\ref{prop:biasvar_main} (variance reduction with $K$), and the complementary roles of clipping and KL penalty.
\end{itemize}

\subsection{Compute Resources}
\label{app:compute}
All experiments were run on a single NVIDIA GeForce RTX 4090 GPU with 24GB of memory. Each training run took approximately 4 to 8 hours, depending on the model size, batch size and dataset, with peak GPU memory usage between 18 and 23 GB. The total compute for the reported experiments is approximately 100 GPU-hours, including preliminary and ablation runs that did not appear in the paper. The full project used approximately 300 GPU hours.

\end{document}